# New Ideas for Brain Modelling 4

Kieran Greer
Distributed Computing Systems, Belfast, UK.
http://distributedcomputingsystems.co.uk

Version 1.2

***Abstract:*** This paper continues the research that considers a new cognitive model based strongly on the human brain. In particular, it considers the neural binding structure of an earlier paper. It also describes some new methods in the areas of image processing and behaviour simulation. The work is all based on earlier research by the author and the new additions are intended to fit in with the overall design. For image processing, a grid-like structure is used with 'full linking'. Each cell in the classifier grid stores a list of all other cells it gets associated with and this is used as the learned image that new input is compared to. For the behaviour metric, a new prediction equation is suggested, as part of a simulation, that uses feedback and history to dynamically determine its course of action. While the new methods are from widely different topics, both can be compared with the binary-analog type of interface that is the main focus of the paper. It is suggested that the simplest of linking between a tree and ensemble can explain neural binding and variable signal strengths.

## 1   Introduction

This paper continues the research that considers a new cognitive model based strongly on the human brain, last updated in [7]. In particular, it considers figure 4 of that paper (Figure 3 below) and how it might be useful in practice. The paper also describes some new methods in the areas of image processing and behaviour simulation. The image processing introduces a most classical form of pattern cross-referencing, while the behaviour equations used feedback for a memory-type of cross-referencing. The work is all based on earlier research by the author and the new additions are intended to fit in with the overall design. For image processing, a grid-like structure is used with 'full linking', if you like. Each cell in the classifier grid stores a list of all other cells it gets associated with and this is used as the learned image that new input is compared with. For the behaviour metric, a new prediction equation is suggested, as part of a simulation, that uses feedback and history to dynamically determine its current state and course of action. While the new methods are from widely different topics, both can be compared with the binary-analog type of interface that is the main focus of the paper. Sensory input may be static and binary, but cross-references result in variable comparisons that make the input more dynamic. It is suggested that the simplest of linking between a tree and ensemble can explain neural binding and variable signal strengths.

The rest of the paper is organised as follows: section 2 introduces an image processing method that cross-references at a pixel level to associate images. Section 3 describes some related work. Section 4 re-visits the behaviour metric of an earlier paper and updates that with a new predictive equation. This feeds earlier evaluations back into the equation, to allow it to self-adjust. Section 6 describes concept aggregation or binding, for realising global concepts, while section 6 considers a process for neural binding that could relate to consciousness. Finally, section 7 gives some conclusions to the work.

## 2   Image Processing

This section describes a very basic image recognition algorithm, but one that has characteristics of the other algorithms developed as part of the work, see the related work section. For this paper, images are represented by a 2-D grid, with a black cell meaning that a pixel is present and a white cell meaning that it is empty. A classifier can use the same grid-like structure, where all of the cells can link to each other. The author has used this structure before in [7] and it is a type of entropy classifier. It attempts to reduce the error overall and is not so concerned with minimising individual associations. The paper [6] describes a classifier that is conjecturally more visual in nature than other types and it also uses a complete linking method. Instead of several levels of feature refactoring, it is a 1-level impression only. With the image classifier, each cell stores a count of every other cell it gets associated with, when averaging this can determine what cells are most similar to the pixel in question. Figure 1 is an example of the clustering technique. If the top LHS grid is the first image to be mapped, then for cell A1, the other black cells are recorded as shown, with a count of 1. The count would then be incremented each time a cell is recorded again, for example, after the second image, cell A3 would lose a count. The idea of linking everything this way has now been used 3 times.

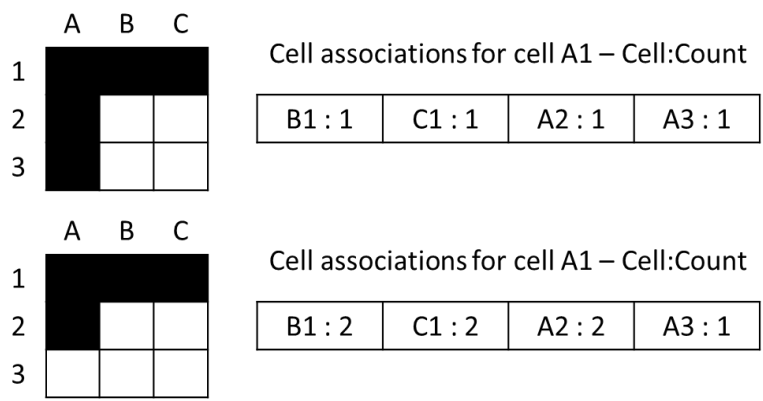

Figure 1. Example mapping of cells presented as an image.

Using this algorithm, a set of hand-written numbers [24] was selected as the test data. There were 9 numbers in total and 55 examples of each number. Each number was trained on a separate classifier, where each cell would store the other related cell associations. The counts could then be averaged to produce the weight values. To use the system, a new binary image would be presented to each of the trained classifiers and it would be assigned to the classifier that matched closest. It is easy to recognise pixels in the input image, but the problem is sorting any other pixels that they are associated with. For each pixel in the input image therefore, the weighted value of the related classifier cell can be retrieved. This would also have links to other cells, maybe not in the input image. For example, if the counts are as shown and a new image is presented that contains pixels in cells A1 and A2 – then the classifier would return cells A1, A2 plus B1 and C1. A3 might not be returned depending on a set threshold value. The success score is then the percentage of retrieved weighted cells in the classifier that are also in the input image, compared to the number that are not in the image. For this example, 2 cells are in the input image while 2 cells are not, leading to a success score of 1. The weight value of the cell can be considered as an association strength and the idea may be auto-associative.

After training on the hand-written numbers, the same dataset was fed through the classifiers again for recognition only. The test results are not particularly good and to improve it, some level of scaling would be required. There is a problem of a larger image covering a smaller one with the currently tried dataset. An average success score of only 46% was achieved with this basic version, with a best score for a number of 89% and a worst score of 15%; although state-of-the-art is only about 55%. It was interesting that the classifier would try to return a picture that was more a reflection of itself. So regardless of what the input was, for example, the number 1 classifier would try to return an image that looked like a '1'. If the input image was a '4' however, then maybe the number 4 classifier would return a more accurate comparison and therefore win the matching competition.

## 3   Related Work

The author's own papers that are quoted [6]-[11] are all relevant to the research of this paper. The image processing of section 2 has already been tried in [3], where they tested the full dataset. Their results were better overall, with maybe 55% accuracy and over a larger dataset. As stated however, the tests here are only initial results and it would be expected that some improvement would be possible, especially if the images can be scaled. The recently found paper [16] looks significant and the general architecture ([7], figure 2, for example)) could have analogies with bi-directional searches in the and-or with theorem-proving graphs architecture of that paper. As suggested, and-or could work from goals to axioms (the neural network in the general model and theorem-proving from axioms to goals (the concept trees in the general model). The paper [23] models at a higher behaviour level, but it is interesting that the behaviours are considered to be unique (time or sequence-based) sets of events and these event patterns are then clustered, rather than each individual event. The idea of using unique sets of nodes to cluster with has also been used for the symbolic neural network [11].

An earlier paper on control theory [25] posts some interesting equations that are similar to ones in this paper. Equation 1 there, for example, looks like the image success score ratio and equation 7 is a likelihood ratio test that is also trying to maximise inside some type of sequence. The behaviour metric of section 4 has only been updated with a new predictive equation, where the first paper [9] notes some references, including [1][12][17] and [19]. One interesting aspect of the equation is that it uses a feedback mechanism which appears to be similar to one that was part of another research project and even in the project code[1].

### 3.1   Cognitive Modelling

Hawkins and Blakeslee [13] describe how a region of the cortex might work (p. 57) and they note an input signal being voted on by a higher level, where one higher level pattern set will win and switch off the other sets. They also state explicitly that the higher level is voting to 'fit' its label better than the other patterns. It may be trying to return its own image as the input signal and the best match there with the input signal should win. The theory that they state is that a region learns when it may be important and then it can become partially active, as part of a memory or prediction. So, this is a type of reasoning, to play over previous scenarios, even when they have not happened in the current situation yet. That can then maybe be reinforced further by specific instance values, making it the real decision. The following quote is also interesting:

'Every moment in your waking life, each region of your neocortex is comparing a set of expected columns driven from above with the set of observed columns driven from below.

---

[1] Cognitive Algorithm by Boris Kazachenko, http://www.cognitivealgorithm.info/.

Where the two sets intersect is what we perceive. If we had perfect input from below and perfect predictions, then the set of perceived columns would always be contained in the set of predicted columns. We often don't have such agreement. The method of combining partial prediction with partial input resolves ambiguous input, it fills in missing pieces of information, and it decides between alternative views.'

The paper [21] introduces the idea of temporal synchrony and synchronised oscillatory activity as important for multisensory perception.

### 3.2   Neural Binding

There is quite a lot of research and philosophy into the idea of neural binding. At its most basic, it means 'how do neural ensembles that fire together be understood to represent the said concept'. For example, questions like 'why don't we confuse a red circle and a blue square with a blue circle and a red square' [4] need to be answered. It includes the idea of consciousness and how the brain is able to be coherent, but while there are lots of theories, there are not a lot of very specific results for the binding mechanism itself. Some cognitive models for the real brain include temporal logic or predicate calculus rules [4] to explain how variables can bind with each other and reasoning can be obtained. This includes the flow of information in both directions and so the basic circuits of this and earlier papers would not be too extravagant. The paper Mashour (2004) is a philosophical paper about how the neural binding mechanism may work. It argues for quantum mechanics, to allow neurons to be represented in more than 1 pattern simultaneously and probably the resulting merging of the patterns into a consciousness. The author would only favour quantum mechanics as a last resort and in section 6, a relatively simple method for representing the same neuron in different patterns is suggested. If time differences between the patterns is very small, then they could still merge into a single coherent message. This could be especially true for the argument against Hebbian cell assemblies. The paper [2] describes a theory that is quite similar. They call the framework the Specialized Neural Regions for Global Efficiency (SNRGE) framework. The paper describes that 'the specializations associated with different brain areas represent computational trade-offs that are inherent in the neurobiological implementation of cognitive processes. That is, the trade-offs are a direct consequence of what computational processes can be easily implemented in the underlying biology.' The specializations of the paper correspond anatomically to the hippocampus (HC), the prefrontal cortex (PFC), and all of neocortex that is posterior to prefrontal cortex (posterior cortex, PC). Essentially, prefrontal cortex and the hippocampus appear to serve as memory areas that dynamically and interactively support the computation that is being performed by posterior brain areas. The PC stores overlapping distributed representations used to encode semantic and perceptual information. The HC stores sparse, pattern separated representations used to rapidly encode ('bind') entire patterns of information across cortex while minimizing interference. The FC stores isolated stripes (columns) of neurons capable of sustained firing (i.e., active maintenance or working memory). They argue against temporal synchrony, because of the 'red circle blue square' question and prefer to argue for coarse-coded distributed representations (CCDR) ([14] and others) instead.

## 4   Cognitive Behaviour

An earlier paper introduced a set of equations that were based on the collective behaviour research of [5]. They proposed a set of characteristics for modelling the stigmergic behaviour of very simple animals, such as ants. They proposed to use coordination, cooperation, deliberation and collaboration, as follows:

- *Coordination* – is the appropriate organisation in space and time of the tasks required to solve a specific problem.
- *Cooperation* – occurs when individuals achieve together a task that could not be done by a single one.
- *Deliberation* – refers to mechanisms that occur when a colony faces several opportunities. These mechanisms result in a collective choice for at least one of the opportunities.

- *Collaboration* – means that different activities are performed simultaneously by groups of specialised individuals.

They note that these are not mutually exclusive, but rather contribute together to accomplish the various collective tasks of the colony. This led to a set of equations by the author [9] for modelling these types of entity. The model is actually behaviour-based not entity-based, where the entity instances are then made up of a set of the pre-defined behaviours, with the following characteristics:

1. *Individual agent characteristics*: Relate to an agent as an individual:
   1.1. *Ability*: this defines how well the behaviour is able to execute the required action.
   1.2. *Flexibility*: this defines how well an agent performing a behaviour can adapt or change to a different behaviour if the situation requires it to. This can be seen as the ability to make that decision individually. The collective capabilities described next can then be seen as the ability to be flexible after an environment response.
2. *Collaborative agent characteristics*: These relate to the agent working in a team environment:
   2.1. *Coordination*: this defines how well the agent can coordinate its actions with those of other agents. This is again a behaviour selection, related to flexibility, but this variable measures the group aspect of the attribute after interaction with other agents.
   2.2. *Cooperation*: this defines how well an agent performing an action can cooperate with other agents also involved in that action. How well can the selected behaviours work together?
   2.3. *Communication*: this defines how well the agents can communicate with each other. This is defined as an input signal and an output signal for each behaviour type. Behaviours could require local or remote communication, for example.

The metric is quite well balanced, with approximately half of the evaluation going to the individual capabilities and half going to the group capabilities.

### 4.1    Problem Modelling

It is possible to specify a problem with all of the related agents and actions that are part of the solution space. The modelling is based around the behaviour types that are used to solve the problem, where the same type definitions can be used, both to model the problem and also to simulate its execution. The agents are defined by agent types, where an agent type can perform a particular set of behaviours. So, if the same behaviour type is to be performed at different levels of success; for static values, this would require different behaviour definitions, or for dynamic ones the value can change through an equation.

### 4.2    Behaviour Equations

The problem is therefore modelled as sets of agents that can each perform a set of behaviours. The Problem Success Likelihood (PSL) is the summed result of the behaviour scores and estimates how well the problem can be solved. The top part of the PSL value, shown in equation 1, evaluates the average agent complexity ($EC_s$), as just described. When modelling, this is measured for all of the behaviour type instances ($B_s$) that are part of the problem behaviour set (PBS). This can be no larger than the optimal problem complexity (PC) value of 1.0. The problem complexity is a factor of how intelligent the agents need to be to solve it. Because the evaluations are all normalised, in a static specification, the maximum value that the problem complexity can be is 1.0. If all behaviours are perfect, they will also only sum to 1 as well. The problem success likelihood, can therefore be defined as follows:

$$PSL = \frac{\left(\sum_{s=1}^{n} EC_s\right)/n}{PC} \quad \forall \, B_s \in PBS$$

Eq. 1

### 4.2.1    Individual Parameters

The individual capabilities of a behaviour can be modelled as follows:

$$EC_s = \frac{I_s + COL_s}{2} \qquad\qquad\qquad \text{Eq. 2}$$

$$I_s = \frac{BA_s + BF_s}{2} \qquad\qquad\qquad \text{Eq. 3}$$

The agent or entity complexity ($EC_s$) for behaviour *s* is a factor of its ability to perform the related behaviour attributes of intelligence and collective capabilities. The agent intelligence ($I_s$) is a factor of its ability ($BA_s$) and flexibility ($BF_s$) capabilities for the specified behaviour.

### 4.2.2   Team Work

The collective or team work capabilities ($COL_s$) of a behaviour are modelled as follows:

$$COL_s = \frac{(COR_s + COP_s + COM_s)}{3} \qquad\qquad\qquad \text{Eq. 4}$$

$$COM_s = \frac{SI_s + SO_s}{2} \qquad\qquad\qquad \text{Eq. 5}$$

The collective capabilities of the agent performing the behaviour are its ability to cooperate with other agents ($COP_s$), coordinate its actions with them ($COR_s$) and also communicate this ($COM_s$), normalised. The communication capabilities of the agent for the behaviour *s* include its ability to send a signal to another agent ($SO_s$) and also its ability to receive a signal from another agent ($SI_s$).

### 4.3   Prediction Operation for the Metric

When simulating the problem, the Problem Complexity value can change. The success likelihood then becomes an individual evaluation, based on its knowledge and understanding of the environment. This can be defined by the subset of behaviours the agent has either performed or has interacted with from other agents. For a more intelligent agent, the memory or history of earlier events can lead to a prediction operation that can reason over the earlier events. It could be a deliberation function that is fed the history of earlier and/or possible choices, before selecting the most appropriate one.

For example, Equation 3 of section 4.2.1 defines agent intelligence as a combination of ability and flexibility. The idea is the ability to perform the intended behaviour but also flexibility to change or adapt the individual behaviour depending on some response. Ideally, an agent would score high in both and a modelling scenario that uses static values would be able to demonstrate this. If instead, running the agents in a simulation, it may be more interesting to let them change their behaviours dynamically, but again constrained by the pre-defined environment. This leads to the idea of a prediction metric that is influenced by what the agent can do and also what it did in the past. The current situation is the most important and so the decision there has the largest weight. The prediction could then include decreasing values for earlier related events. These can be factored as a count of earlier events times a factor for the time when they occurred. If the behaviour was not repeated, then maybe something went wrong, such as an unfavourable response. These responses, including for the current situation, would change the state of the agent into what it then has to deal with. The equation would be something like:

$$\mathrm{Pr} = (\mathrm{EC}_{S1} + (\textstyle\sum_{m=0}^{M} f(\mathrm{n}, \mathrm{EC}m, \mathrm{R}, \mathrm{t}) \,/\, M)) \,/\, 2 \qquad\qquad \text{Eq. 6}$$

Where:

$EC_{s1}$ is the currently selected individual behaviour complexity,

$EC_m$ is any previously selected individual behaviour complexity, for any related scenario,

n is number of times in memory that the previous event occurred,

R is the response or impact of the event,

t is the last time the event occurred,

M is the total number of behaviour-response pairs stored in memory,

$f$ is some function evaluation over the variable set, maybe 'n(EC + R) / t' if R is a specific response, or a multiplication if it is the weight of the response.

When the agent selects a new behaviour, it is expecting a positive response. After a reply from the environment, the individual behaviour plus the response is fed back into the equation to get a new amount. If this is less than expected – 'current Pr plus new EC', then the response has been a negative one and possibly a different behaviour should be selected. This type of process can repeat, with bad responses being flagged and not selected again, for example, until a decision is made, maybe a new stable state is reached. As the equation calculates, it also feeds back its current state to update its evaluation for the next selection. The responses are therefore even more responsible for changing the agent state, where the behaviour selection, using the entity complexity, is an individual one, maybe based more on knowledge. So again, there is a hint of a simpler evaluation, which is the knowledge-based decision, balancing itself with the more complex decision, after the response is also factored in.

## 4.4   Self-Adjusting Evaluation

A more intelligent version of the metric for a simulation might therefore look as follows:

$$PSL = \frac{Pr}{PC} \quad \forall\, Bs\, \in PBS \qquad\qquad\qquad\qquad\qquad \text{Eq. 7}$$

Where the predictive equation can replace the entity complexity and both of the flexibility parts. The prediction is modelled as in equation 6, by the agent decision plus its reaction to any response. A dynamic problem complexity can be measured as in equation 8 and is essentially the fraction of all behaviours that the agent knows about. Depending on how this is measured, a multiplication might be more appropriate for the PSL. However, if I only 'know' about 1 behaviour, for example, then based on my own knowledge, I can solve that more easily than if I have to deal with several known behaviours.

$$PC = \left(\sum_{n=0}^{N} ECn + \sum_{m=0}^{M} Rm\right) / PBS \qquad \text{Eq. 8}$$

The collective capabilities are thus reduced to cooperation and communication with other entities, where coordination is moved to the predictive part.

$$COL_s = \frac{(CORs + COPs + COMs)}{3}\frac{COPs + COMs}{2} \qquad \text{Eq. 9}$$

### 4.5    Testing

The behaviour metric was tested in [9]. There has not been an opportunity to test the new predictive equation or its feedback algorithm and so this paper presents the theory of it only and notes the relation of the theory to the other research. However, a worked example described next, should help to show how it would work in practice.

### 4.5.1    Worked Behaviour Example

Consider the following scenario: an agent finds itself in a situation $S_1$. The agent is modelled with behaviours $B_1$ to $B_5$ and the world is modelled with behaviours $B_1$ to $B_{10}$. The agent has encountered the same scenario $S_1$ before and attempted the following behaviour set with related events, to deal with it:

Event $B_3$ was tried at time $t_3$ and resulted in a response $R_8$.
Event $B_4$ was tried at time $t_2$ and resulted in response $R_6$.

Based on this, behaviour $B_4$ is selected again, leading to the equation:

$Pr = B_{4S1} + (1 \times (B_4 + R_6) / 2) + (1 \times (B_3 + R_8) / 3)$

The scene is an interactive one, with another agent able to reply. The other agent also knows the scenario and replies with a behaviour B10. This is found to be unfavourable for the agent and reduces its overall evaluation through the equation:

$Pr = (B_{4S1} + R_{10S1}) + (1 \times (B_4 + R_6) / t_2) + (1 \times (B_3 + R_8) / t_3)$, where $R_{10S1}$ is negative.

Therefore, the prediction reduces and the agent is required to try again. It now has knowledge of the reply B10 and also knows not to use behaviour B4 if it doesn't have to. Therefore, a new response based on its new history could lead to:

$Pr = B_{2S1} + (1 \times (B_{4S1} + R_{10S1}) / t_2) + (1 \times (B_4 + R_6) / t_3) + (1 \times (B_3 + R_8) / t_4)$

The reply by the environment, $B_{10}$ again, is not as unfavourable now and so the prediction increases. Based on other criteria, the behaviour can be played again, or a satisfactory situation may have been achieved. In either case, to resolve this situation, two behaviours were tried and both were fed back into the evaluation function. The first one even counted as part of the history for selecting the second one.

## 5   Concept Binding

Concept aggregation or binding is mentioned as part of the symbolic neural network [11]. This is an experience-based structure that combines lower-level concepts into more complex global ones and would work in the more intelligent brain region. Other related papers [6][7] have developed algorithms that link everything together and then sort using entropy. The image-processing model of section 2, for example, introduces the variable structure through cross-referencing. The behaviour metric of section 4 does not fit quite as obviously. The sensory input must still come first from the environment, before deciding on a plan. The behaviour selection and reasoning process must therefore occur afterwards, using the higher-level brain regions to sort the lower-level ones. But a binding is simply a repeat of the pattern and it does not have to represent anything other than what the ensemble mass represents. So it can be used in exactly the same context. With one visual system theory ([22] and others), synchronous oscillations in neuronal ensembles bind neurons representing different features of an object. Gestalt psychology is also used, where objects are seen independently of their separate pieces. They have an 'other' interpretation of the sub-features and not just a summed whole of them. Although, there are still problems with the theories, including the requirement for too many independent neural constellations to represent every feature.

The numerical problem can therefore be helped with some level of cross-referencing. The question like 'why don't we confuse a red circle and a blue square with a blue circle and a red square' [4] could be answered if 'red', 'blue', 'circle' or 'square' are individual concepts that also cross-reference each other. Individual means a base node in a tree and cross-reference means a leaf node in another tree. If a tree is accepted as part of a circuit, then the base neuron will receive positive feedback, which may be recognised more because of the greater firing effort. Leaf nodes would also have to be relevant to complete a circuit, but they may also be peripheral to a main concept and so can act more as links. One leaf node would also be the base of another tree, when both trees could be active at the same time and relate to the same larger concept. This is illustrated in Figure 2. The concepts of red, blue, circle and square are all base concepts learned by the system and also have cross-referenced branches in other trees. Some neurons can have 1000 branches or more[2]. If the senses send signals about red and circle, for example, then the two central trees can complete a circuit, even if the concepts exist in other places as well. For reasoning then, there would also be influences from the higher-level processes that perform other types of aggregation, or for synchronisation.

---

[2] Prof K. Arai, SAI'14.

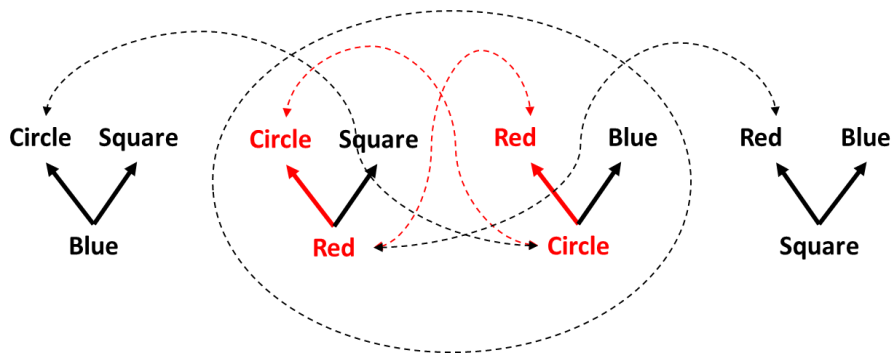

Figure 2. One level of linking in a temporal model defines a particular ensemble mix.

It is therefore possible to give concepts graded strengths and also any arbitrary mix of the learned base set. This is already part of the concept trees research [8], where a leaf node in one tree links to a base node in another tree. Or to put it another way, if a base node branches to link with another tree, it represents the same thing in any other tree. The concept tree may be too semantic for the level being considered here and their base concepts would not be 'anchored' because there is an idea that concept trees can re-join with each other. So possibly, the path description for Figure 2 is only 1 or 2 levels deep – the base node and its same branches. The related Concept Base [11] however has been used to manage flat hierarchies in that paper, or the trees in other papers, where the flat hierarchy has been associated with a cognitive process.

## 6   Neural Binding for a Binary-Analog Interface

While section 5 looked at the coarser concept binding, this section considers again the idea of neuron binding[3]. Biology has already suggested theories about neural oscillations and binding that include neural pairing. The pairing helps to group the neurons into specific patterns that the brain can understand, when different features can become synchronised and oscillate together.

---

[3] See Wikipedia, for example.

The binding theory of this paper is described in Figure 3. In the figure, a neuron in a base ensemble mass binds with a neuron in a related hierarchical structure. The hierarchy gives more meaning to the structure and helps to guide a search process, but it is not clear how or why this structure would form. Signal strength would be an attractive option to produce the hierarchical neuron, but if the base node has the strongest signal, then that should result in a longer link to its paired neuron. The construction of the hierarchy is therefore more likely to be based on time. When neurons fire they create new neurons and a neuron must exist before it can form a link to another one. The neurons that form first are therefore more likely to link to other neurons that form later and so in a mechanical sense, the hierarchy could be created.

It is also useful to consider electrical synapses, which can be bi-directional and setup an oscillating wave between close neural regions. They are created along with chemical synapses. The point of the pairing is this resonance and so a weaker electrical signal would be ideal. Quantum mechanics is one theory used to exaplain how the conscious might work, where several patterns and states can collapse into one. With a paired neuron however, there can be resonance between the pair without a quantum element. This resonance could produce a signal, similar to how different sized pipes produce a note. The resonance is obviously very quick and so it would all meld into the one signal. If the base ensemble can refresh the whole pattern as well, then that variable process can (re)activate parts of the structure and in a timed way. This model fits deeper in the brain however and is not intended for the intelligent cortex area.

The model is also based on the idea of an auto-associative neural network. The Hopfield neural network [15], and its stochastic equivalents are auto-associative or memory networks. With the memory networks, information is sent between the input and the output until a stable state is reached, when the information does not then change. These are resonance networks, such as bidirectional associative memory (BAM), or others [20], but they can only provide a memory recall – they map the input pattern directly to the output pattern. If some of the input pattern is missing however, they can still provide an accurate recall of the whole pattern. They also prefer the data vectors to be orthogonal without overlap. This is however ideal for the binding that only wants to reproduce the base ensemble in the hierarchy.

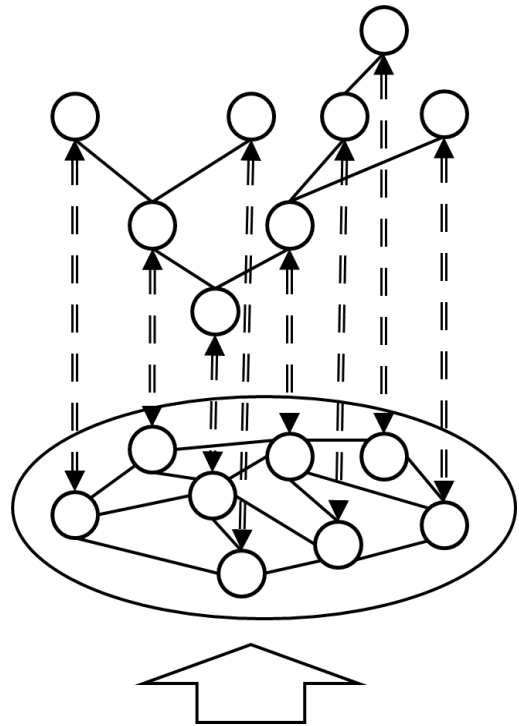

Figure 3. Neuron Pairing: an ensemble neuron links with a hierarchal neuron. Also figure 4 in Greer (2016).

## 6.1    Image Processing Example

An an example, the image-processing algorithm of section 2 is mapped to the neuron pairing architecture in this section. This is not final and there are questions about how exactly it might work, but there is also a clear process that can relate the two. The first thing to consider is the neuron ensemble and while neurons are continually being created, the ensemble mass is assumed to exist already. When some of it is excited, this then starts the binding process with the neurons in the hierarchy. The process is shown in Figure 4. The LHS represents the lower ensemble mass, where an input has activated the central column of black neurons. With the image-processing algorithm, each pixel (neuron) on the RHS image relates to the same pixel (neuron) on the LHS image. The grey squares represent additional pixels (neurons) that have been associated during the reinforcement procedure. The binding process can probably include more than 1 base beuron and hierarchy at a time and it is probably not the case that only the grey squares would be further up a hierarchy. While that part is not clear, what is good about the binding is that it should introduce more accuracy into the recognition result. It would require for both the input sensory image and the stored hierarchy image to both fire the same related set of neurons, for the oscillations to register a persistent signal between them. If there is a neuron in the hierarchy that sends a signal back to the sensory input and it is not part of the pattern, then it cannot oscillate, so that part of the error can be removed. If part of the sensory input is missing from the hierarchy, then it cannot oscillate and so that part of the input needs to be learned.

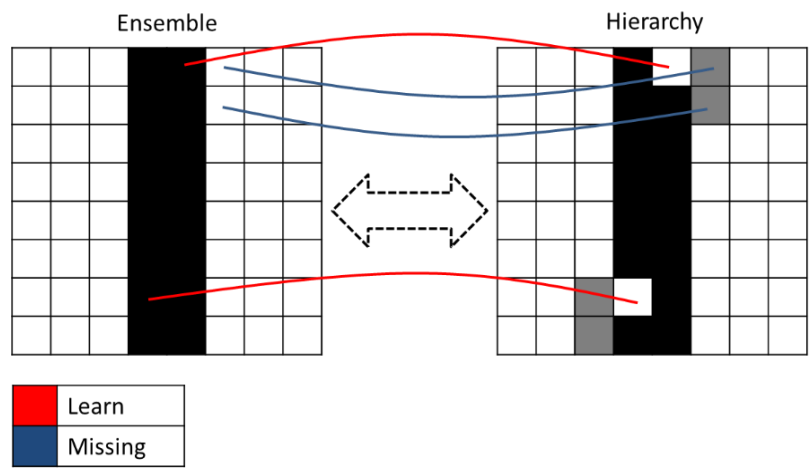

Figure 4. LHS relates to neuron binding ensemble mass, with central column activated. RHS relates to hierarchy, with a direct mapping. The two red lines show where the ensemble is missing and so it needs to be learned. The blue lines show extra neurons from the hierarchy back to the ensemble, but can be removed as error. The other paired black squares represent where the patterns match and can oscillate together.

A third possibility is if the hierarchy returns a signal not in the sensory input, but that has links in the ensemble mass. It may usually be part of the input pattern and so through the links it can get activated as part of the pattern. A fourth possibility is if a lot of the hierarchy sends back signals to inactive neurons, but the hierarchy should be more accurate and is activated from the ensemble. It is also controlled from further levels above, so this possibility is not as likely.

## 7    Conclusions

This paper has mostly considered a binding problem and the implications of a binary-analog conversion process. This has led to some interesting results, both in the areas of image processing and higher-level reasoning. It has also helped when thinking about the temporal synchrony problems of neural binding and how separate parts can be re-combined into a more coherent whole. The binding can also help with recognition accuracy. The behaviour metric has been updated to include a predictive part that may be used as part of a simulation. This allows the metric to be more intelligent and should help to clarify the flexibility attributes that it has. A behaviour decision is based on the agent's current state, its abilities and also its memory.

If the model of this and earlier papers is used, then (sub)concepts can in fact be represented individually, with lots of cross-linking representing the different contexts. With so many neurons in the brain, depending on how a scenario is broken down, why could it not accommodate this? For the pattern ensembles, base nodes that link as branches in other trees can simply represent themselves. This is a simplification of concept trees, where symbolically or conceptually, the node representation is only 1 or 2 levels deep. If there is a tree structure however, then there can be deeper paths that can relate to graded signal strengths, or even allow for different connection patterns over the same ensemble. Also key is reinforcement from above, through reasoning, that completes the circuits. It can still be shown that the ideas fit together into a common model, even if it uses a lot of standard and compatible structures.